\documentclass[10pt,twocolumn,letterpaper]{article}

\usepackage{cvpr}

\usepackage{graphicx}
\usepackage{amsmath}
\usepackage{amssymb}
\usepackage{booktabs}
\usepackage{ctable}
\usepackage{multirow}
\usepackage{array}
\usepackage{colortbl}
\usepackage{float}
\usepackage{stfloats}

\definecolor{cvprblue}{rgb}{0.21,0.49,0.74}
\usepackage[pagebackref,breaklinks,colorlinks,citecolor=cvprblue]{hyperref}

\def\paperID{23}
\def\confName{Mobile AI Workshop}
\def\confYear{2026}

\begin{document}
	
\title{Real Image Denoising with Knowledge Distillation\\ for High-Performance Mobile NPUs}

	\author{Faraz Kayani{\thanks{Corresponding author: faraz.kayani2322@gmail.com}}, \space\space\space Sarmad Kayani,\space\space\space Asad Ahmed,\space\space\space Radu Timofte,\space\space\space Dmitry Ignatov{\thanks{ dmytro.ignatov@uni-wuerzburg.de}} \\
		\small{Computer Vision Lab, CAIDAS \& IFI, University of W\"urzburg, Germany}}
	\maketitle

\begin{abstract}

While deep-learning-based image restoration has achieved unprecedented fidelity, deployment on mobile Neural Processing Units (NPUs) remains bottlenecked by operator incompatibility and memory-access overhead. We propose an NPU-aware hardware--algorithm co-design approach for real-world image denoising on mobile NPUs. Our approach employs a high-capacity teacher to supervise a lightweight student network specifically designed to leverage the tiled-memory architectures of modern mobile SoCs. By prioritizing NPU-native primitives---standard $3\times3$ convolutions, ReLU activations, and nearest-neighbor upsampling---and employing a progressive context expansion strategy (up to $1024\times1024$ crops), the model achieves 37.66~dB PSNR / 0.9278 SSIM on the validation benchmark and 37.58~dB PSNR / 0.9098 SSIM on the held-out test benchmark at full resolution ($2432\times3200$) in the Mobile AI 2026 challenge. Following the official challenge rules, the inference runtime is measured under a standardized Full HD ($1088\times1920$) protocol, where it runs in 34.0~ms on the MediaTek Dimensity 9500 and 46.1~ms on the Qualcomm Snapdragon 8 Elite NPU. We further reveal an ``Inference Inversion'' effect, where strict adherence to NPU-compatible operations enables dedicated NPU execution up to $3.88\times$ faster than the integrated mobile GPU. The 1.96M-parameter student recovers 99.8\% of the teacher's restoration quality via high-$\alpha$ knowledge distillation ($\alpha=0.9$), achieving a $21.2\times$ parameter reduction while closing the PSNR gap from 1.63~dB to only 0.05~dB. These results establish hardware-aware distillation as an effective strategy for unifying high-fidelity denoising with practical deployment across diverse mobile NPU architectures. The proposed lightweight student model (\textit{LiteDenoiseNet}) and its training statistics are provided in the NN Dataset, available at \href{https://github.com/ABrain-One/NN-Dataset}{https://github.com/ABrain-One/NN-Dataset}.
\end{abstract}

\section{Introduction}
\label{sec:intro}

Real-world image denoising is a core component of mobile computational photography. Deep neural networks have greatly improved restoration quality, moving from synthetic Gaussian noise removal to suppression of complex, spatially variant sensor noise in smartphone images. However, a substantial deployment gap remains between high-performing desktop architectures and practical mobile solutions, where computation, memory footprint, and thermal stability are tightly constrained.

A major obstacle in mobile AI deployment is the mismatch between desktop-oriented restoration models and mobile Neural Processing Units (NPUs). Many high-performing methods rely on global self-attention, deformable convolutions, or complex dynamic upsampling operators. Although effective on desktop hardware, these components often lack efficient native support on mobile NPUs or DSPs, causing framework-level CPU fallbacks and undermining real-time deployment.

In this work, developed in the context of the Mobile AI 2026 Image Denoising benchmark, we study real-image denoising under strict mobile deployment constraints. Rather than adapting a heavy restoration network after training, we constrain the student architecture from the outset to use hardware-friendly operators suited to native NPU execution. A high-capacity teacher transfers restoration knowledge to this compact student through high-$\alpha$ distillation, while progressive context expansion during fine-tuning improves full-resolution structural recovery.

\begin{figure*}[!t]
    \centering
    \includegraphics[width=\textwidth]{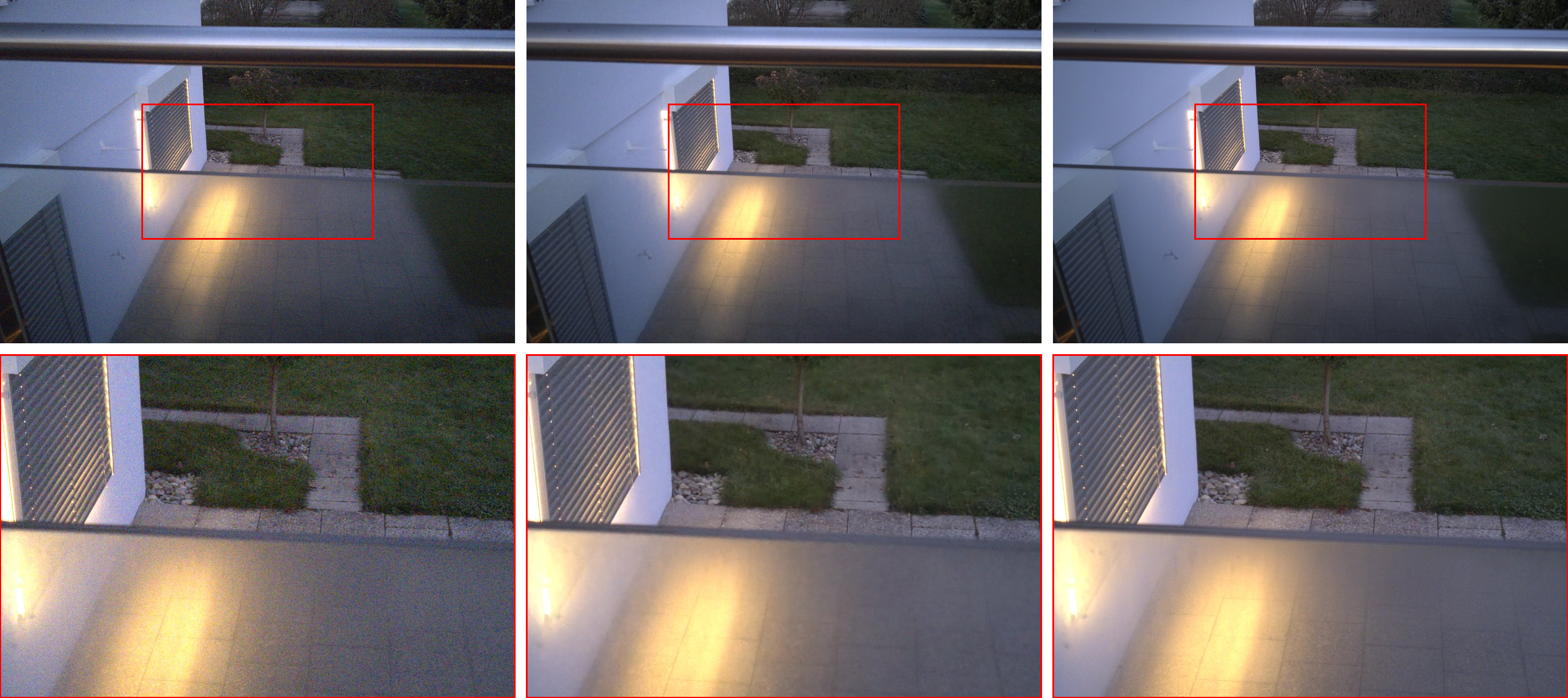}
    \caption{Qualitative comparison on a representative validation example. The first row shows the full images, while the second row presents zoomed crops from the same regions. From left to right: noisy input, restored output of the proposed lightweight student model, and ground-truth clean image. The proposed student effectively suppresses visible noise while preserving local structures and fine textures.}
    \label{fig:teaser}
\end{figure*}

The resulting model attains 37.66 dB PSNR / 0.9278 SSIM on the full-resolution validation benchmark and 37.58 dB PSNR / 0.9098 SSIM on the held-out test benchmark, while running in 34.0 ms on the MediaTek Dimensity 9500 NPU and 46.1 ms on the Qualcomm Snapdragon 8 Elite NPU under the official Full HD protocol. As shown in Figure~\ref{fig:teaser}, the proposed student suppresses visible sensor noise while preserving local structures and fine textures.

In summary, this paper makes four contributions: (1) a hardware-aware teacher--student approach for real-image denoising on modern mobile NPUs; (2) a compact student architecture built entirely from hardware-friendly operators for stable fallback-free execution; (3) high-$\alpha$ distillation with progressive context expansion, allowing a 1.96M-parameter student to recover 99.8\% of the teacher's restoration quality while reducing the validation PSNR gap from 1.63 dB to 0.05 dB; and (4) strong challenge results with official NPU runtimes of 34.0 ms on the Dimensity 9500 and 46.1 ms on the Snapdragon 8 Elite.

\section{Related Work}
\label{sec:Related}

\subsection{Image Denoising and Real-Noise Restoration}
Image denoising has evolved from model-based formulations to deep networks that directly map noisy inputs to clean targets. Early deep denoisers such as DnCNN and FFDNet established convolutional residual learning as a strong baseline for denoising~\cite{zhang2017dncnn,zhang2018ffdnet}. For real-image denoising, CBDNet and RIDNet addressed spatially variant, camera-dependent noise more explicitly~\cite{guo2019cbdnet,anwar2019ridnet}. More recent restoration architectures, including MIRNet, MPRNet, HINet, SwinIR, Uformer, Restormer, and NAFNet, further improved denoising quality through multi-scale feature extraction, progressive refinement, normalization-aware restoration design, and transformer-based global or hierarchical modeling~\cite{zamir2020mirnet,zamir2021mprnet,chen2021hinet,liang2021swinir,wang2022uformer,zamir2022restormer,chen2022nafnet}. Despite their strong performance, many of these models remain relatively heavy for direct mobile deployment.

\subsection{Real-Noise Benchmarks}
Progress in real-image denoising has been strongly influenced by paired noisy/clean datasets and benchmark challenges. DND and SIDD established widely used benchmarks for real-image and smartphone denoising under captured noise conditions~\cite{plotz2017dnd,abdelhamed2018sidd}, while the NTIRE real image denoising challenges further emphasized restoration under realistic degradations~\cite{abdelhamed2019ntire,abdelhamed2020ntire}. More recently, Flepp \textit{et al.} introduced the Mobile Image Denoising Dataset (MIDD), explicitly targeting mobile denoising across diverse smartphone sensors with efficient baselines~\cite{flepp2024midd}. Together, these benchmarks shifted denoising research toward realistic and deployment-relevant evaluation.

\subsection{Knowledge Distillation for Image Restoration}
Knowledge distillation transfers information from a larger teacher model to a smaller student. Hinton \textit{et al.} introduced the general teacher--student distillation paradigm~\cite{hinton2015distilling}, while FitNets demonstrated the benefit of intermediate guidance for thinner and faster students~\cite{romero2015fitnets}. In image restoration, distillation is particularly attractive because lightweight models often suffer from reduced representational capacity. Young \textit{et al.} proposed feature-level distillation for efficient RAW image denoising on resource-constrained devices~\cite{young2022featurealign}, and Li \textit{et al.} explored denoising-specific heterogeneous knowledge transfer in a lightweight denoising framework~\cite{li2022mdrn}. Unlike prior restoration distillation works, our goal is not only to recover accuracy in a compact student, but to distill the teacher into an explicitly NPU-native operator set for stable fallback-free execution.

\subsection{Mobile and Deployment-Aware Image Restoration}
While many restoration methods report strong results on desktop GPUs, actual mobile deployment imposes strict constraints regarding memory bandwidth, supported operators, and thermal envelopes. Practical mobile denoising has therefore motivated a line of work that explicitly trades architectural complexity for device efficiency. Wang \textit{et al.} presented a lightweight raw-image denoising framework designed for practical execution on mainstream mobile devices~\cite{wang2020practical}. Liu \textit{et al.} proposed MFDNet, a mobile-friendly denoising network built through an analysis of operations that execute efficiently on NPUs~\cite{liu2022mfdnet}. Flepp \textit{et al.} introduced efficient baselines for real-world mobile denoising and showed that compact models can directly process high-resolution mobile images while remaining compatible with recent smartphone accelerators~\cite{flepp2024midd}. More generally, lightweight restoration modules and baselines have also been proposed to reduce the computational burden of low-level vision models on resource-constrained hardware~\cite{lahiri2020lightweight,fan2024lir}.

The Mobile AI Workshop series explicitly highlights the deployment gap by evaluating models directly on smartphone SoCs~\cite{ignatov2021mai_denoising,mai2026workshop,mai2026report}. Our work is aligned with this deployment-oriented direction, but differs in emphasizing native NPU execution as a first-class design target. Rather than starting from a high-performing restoration model and adapting it after training, we constrain the student architecture from the outset to use hardware-friendly operators that avoid fallback behavior on dedicated mobile AI accelerators.
	
\section{Methodology}
\label{sec:Methodology}

\begin{figure*}[!t]
    \centering
    \includegraphics[width=\textwidth]{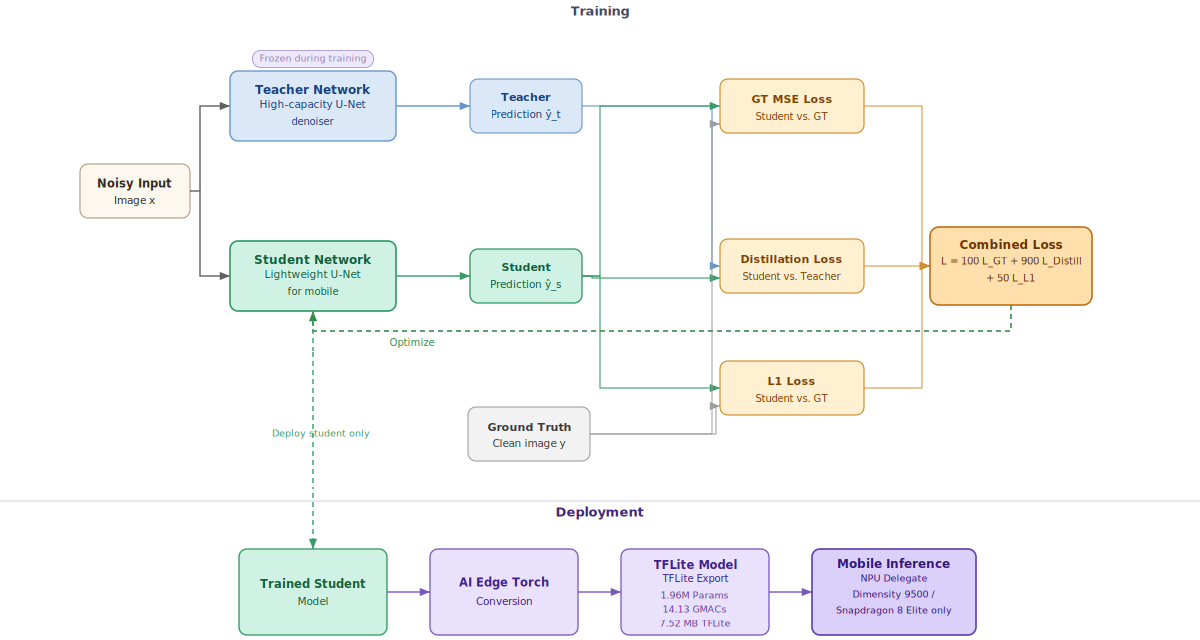}
    \caption{Overview of the proposed teacher--student mobile denoising pipeline. During training, a high-capacity teacher supervises a lightweight student using ground-truth reconstruction, distillation, and L1 losses. During deployment, the trained student is exported through AI Edge Torch to TensorFlow Lite for fallback-free NPU inference.}
    \label{fig:pipeline}
\end{figure*}

We propose a hardware--algorithm co-design approach for real-world image denoising, using high-$\alpha$ knowledge distillation from a high-capacity teacher to a highly constrained, NPU-native student. The goal is to retain strong restoration fidelity while adhering to the operator and memory-access limitations of modern mobile NPUs. The overall training and deployment pipeline is illustrated in Figure~\ref{fig:pipeline}.

\subsection{High-Capacity Teacher Network}
Encoder--decoder networks with skip connections have become a standard design pattern for dense image restoration and related dense prediction tasks~\cite{ronneberger2015unet}. Building on this design family, we employ a high-capacity U-Net-style denoising architecture as the teacher model, designed purely for maximum restoration fidelity.

The teacher follows a three-level encoder--bottleneck--decoder topology (64--512 channels). An \textit{InputBlock} maps RGB input to 64 channels via two $3\times3$ convolutions. Each encoder stage applies two \textit{DenoisingBlocks} and a \textit{DownsampleBlock} ($2\times2$ stride-2 convolution). The bottleneck stage contains two additional \textit{DenoisingBlocks} at the lowest spatial resolution.

The decoder mirrors the encoder. Each decoder stage begins with an \textit{UpsampleBlock}, which first upsamples features using a $2\times2$ transposed convolution, concatenates the corresponding encoder skip feature, and then applies a $3\times3$ convolution with PReLU activation to fuse the combined features. After each upsampling stage, two \textit{DenoisingBlocks} are used for feature refinement. Finally, an \textit{OutputBlock} consisting of two $3\times3$ convolutions maps the 64-channel decoder output back to the image space.

Each \textit{DenoisingBlock} uses four $3\times3$ convolutions with dense internal feature aggregation and a local residual connection. More specifically, given an input with $f$ channels, the block produces three intermediate feature groups of width $f/2$, each concatenated to the running feature tensor, and a final $3\times3$ convolution maps the aggregated features back to $f$ channels. PReLU activations are used after every convolution, and the block output is added to the original input through a residual connection.

Overall, the teacher contains 41.6M trainable parameters and achieves 37.71 dB PSNR on the validation benchmark. It serves as the fixed high-capacity supervisor during student distillation. However, its wide feature maps, transposed convolutions, and large intermediate activations make it poorly suited to mobile deployment, exceeding the effective memory budget of the target NPU at high resolution.

\subsection{Hardware-Aware Student Network}
The deployable student model is an ultra-lightweight encoder-decoder network specifically architected for fallback-free NPU execution. To maximize throughput on the tiled-memory architectures of both the MediaTek APU and Qualcomm Hexagon processors, the student incorporates three strict hardware-aware design principles.

First, we utilize a highly vectorized base width of 16 channels, scaling up progressively across four downsampling stages. This choice aligns well with the SIMD execution characteristics of modern mobile accelerators and improves utilization of vectorized units.

\begin{figure}[h]
    \centering
    \includegraphics[width=0.95\linewidth]{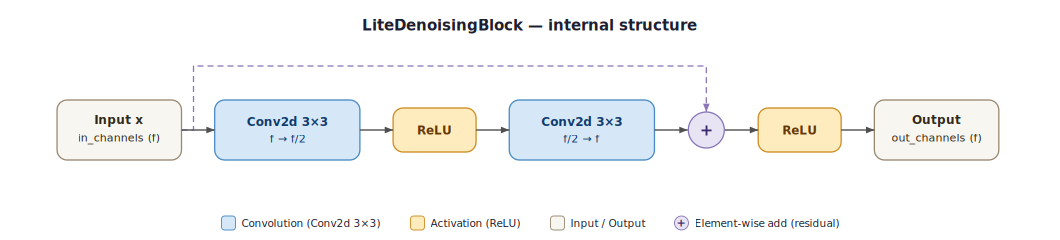}
\caption{Internal structure of the LiteDenoisingBlock used throughout the student network. The block employs an internal channel-reduction bottleneck ($f \rightarrow f/2$) between standard $3\times3$ convolutions, together with hardware-native ReLU activations and a local residual connection.}
    \label{fig:lite_block}    
\end{figure}

Second, the core computational unit of our network is the \textit{LiteDenoisingBlock}, detailed in Figure~\ref{fig:lite_block}. Each block is constrained to standard $3\times3$ convolutions paired strictly with hardware-friendly ReLU activations, avoiding the latency penalties associated with complex non-linearities (e.g., GELU or Swish) on mobile platforms. To reduce memory bandwidth overhead, the block incorporates an internal channel-reduction bottleneck ($f \rightarrow f/2 \rightarrow f$), reducing MACs while preserving a residual connection.

\begin{figure*}[!t]
    \centering
    \includegraphics[width=\textwidth]{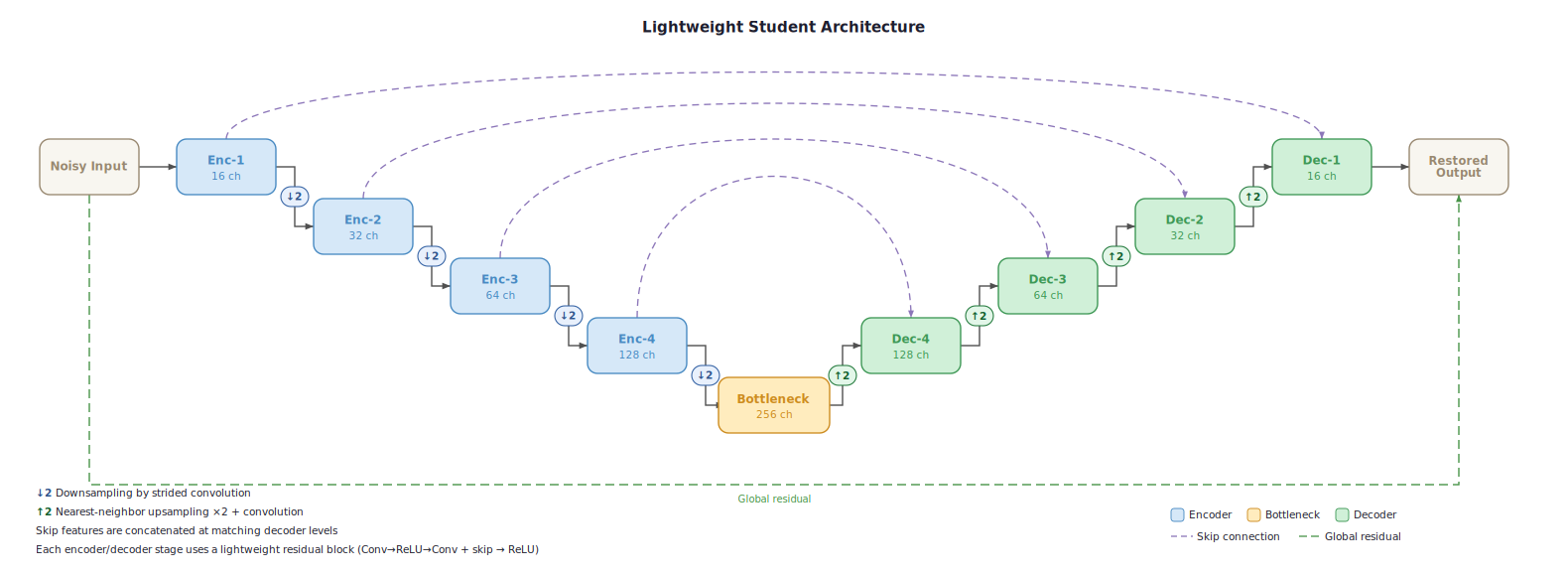}
    \caption{Simplified internal architecture of the lightweight student denoising model. The network follows a compact U-Net-style design~\cite{ronneberger2015unet}. The repeated processing units in the encoder and decoder are LiteDenoisingBlocks, whose internal structure is shown in Figure~\ref{fig:lite_block}. Crucially for NPU deployment, downsampling is performed via strided convolutions, while the decoder uses hardware-friendly nearest-neighbor upsampling followed by convolutional refinement rather than transposed convolutions.}
    \label{fig:student_architecture}
\end{figure*}

Finally, and most crucially, we completely replace the transposed convolutions used in the teacher's decoder with parameter-free nearest-neighbor upsampling followed by a $3\times3$ convolutional refinement. This eliminates the complex zero-padding logic that frequently causes hardware fallbacks on less flexible NPU drivers. As in many restoration networks, the final output is produced through global residual learning:
\begin{equation}
    \hat{I}_s = \mathrm{clip}(f_{\theta}(I_{\mathrm{noisy}}) + I_{\mathrm{noisy}}, 0, 1),
\end{equation}
where $f_{\theta}$ denotes the student network and $\hat{I}_s$ is the restored image. This topology, illustrated in Figure~\ref{fig:student_architecture}, allows the student to remain remarkably compact (1.96M parameters) while perfectly conforming to the fast-path execution graphs of flagship NPUs.

\subsection{Distillation Objective}
To bridge the representational capacity gap between the 1.96M parameter student and the massive teacher, we employ a high-$\alpha$ knowledge distillation strategy. Let $\hat{I}_s$ denote the student prediction, $\hat{I}_t$ the frozen teacher prediction, and $I_{gt}$ the clean target image. The base reconstruction loss against the ground truth is defined as:
\begin{equation}
    \mathcal{L}_{gt} = \|\hat{I}_s - I_{gt}\|_2^2,
\end{equation}
while the distillation loss, which forces the student to emulate the teacher's noise-free continuous manifold, is defined as:
\begin{equation}
    \mathcal{L}_{distill} = \|\hat{I}_s - \hat{I}_t\|_2^2.
\end{equation}
To further enforce local structural fidelity and stabilize high-frequency texture generation, we incorporate an L1 penalty against the ground truth:
\begin{equation}
    \mathcal{L}_{L1} = \|\hat{I}_s - I_{gt}\|_1.
\end{equation}
The final objective function aggressively prioritizes the teacher's guidance to prevent the student from overfitting to residual sensor noise present in the raw ground truth data:
\begin{equation}
    \mathcal{L}_{total} = \lambda_{gt}\mathcal{L}_{gt} + \lambda_{distill}\mathcal{L}_{distill} + \lambda_{L1}\mathcal{L}_{L1},
\end{equation}
where our optimal configuration utilizes $\lambda_{gt}=100$, $\lambda_{distill}=900$, and $\lambda_{L1}=50$.

\subsection{Progressive Context Expansion Strategy}
Training is executed within the \texttt{LEMUR NN Dataset} framework~\cite{ABrain.NN-Dataset}, leveraging Optuna-based hyperparameter exploration. Because the student model possesses a highly restricted receptive field due to its lightweight convolutions, it naturally struggles to resolve broad spatial structures. To counteract this, we introduce a Progressive Context Expansion strategy. 

Models are initialized and trained extensively on $256\times256$ paired crops. During the critical fine-tuning stages, the spatial context is systematically expanded to $512\times512$ and ultimately to $1024\times1024$ crops. Exposing the lightweight model to these massive spatial contexts allows it to capture global luminance gradients and structural coherence that are mathematically invisible at lower resolutions, drastically improving final full-resolution inference quality.

\subsection{Cross-Platform Deployment Pipeline}
The ultimate objective of this approach is fallback-free mobile deployment. The trained student model is exported to the TensorFlow Lite (TFLite) format utilizing AI Edge Torch~\cite{ABrain.NN-Lite}. A critical step in this conversion involves tensor layout permutation. While PyTorch utilizes an NCHW (\textit{Batch, Channel, Height, Width}) memory format, optimized mobile NPU accelerators require an NHWC layout to maximize continuous memory reads during spatial convolutions. We introduce a wrapper during the export phase to natively fuse these permutations into the execution graph, ensuring the model is "plug-and-play" for standard Android image buffers.

To ensure the conversion process preserves the learned restoration manifold, we perform a rigorous numerical parity check between the PyTorch reference and the exported TFLite model executed on the host CPU. This comparison confirms a maximum deviation of less than $10^{-4}$. 

For deployment, two TFLite configurations are generated: a dynamic-resolution model for full-resolution quality evaluation ($2432 \times 3200$) and a fixed-resolution FP16 model with input shape $1 \times 1088 \times 1920 \times 3$ for runtime benchmarking. Mobile evaluation is performed with the AI Benchmark application using native NPU delegates.

\section{Experiments}
\label{sec:experiments}

\subsection{Dataset and Augmentation}
We evaluate the proposed approach on the official Mobile AI 2026 Image Denoising Challenge dataset, which contains 650 paired noisy/clean training images and 50 validation pairs. We apply a 2-pixel buffer padding before extracting $256 \times 256$ crops to reduce edge bias. Geometric augmentation includes random horizontal and vertical flips together with orthogonal rotations ($90^\circ, 180^\circ, 270^\circ$). As part of progressive context expansion, training crops are later increased to $512^2$ and $1024^2$ during fine-tuning.

\subsection{Implementation Details}
The student model is implemented in PyTorch 2.4 and trained on a single NVIDIA RTX 4090 GPU. We use the \texttt{LEMUR NN Dataset} framework for data orchestration and Optuna for hyperparameter optimization. Optimization is performed with Adam and a Cosine Annealing learning rate schedule ($T_{max}=200$, $\eta_{min}=1e-5$), together with gradient clipping at a maximum norm of 0.1 under high-$\alpha$ distillation ($\alpha=0.9$).

The final reported model is trained for 200 epochs and then fine-tuned for an additional 20--30 epochs. Most optimization is performed on $256\times256$ crops, with larger $512\times512$ and $1024\times1024$ crops introduced during the final fine-tuning stage. On a single NVIDIA RTX 4090 GPU, the complete training and fine-tuning pipeline typically requires 3--6 hours depending on batch size and crop configuration. The final model is exported to TensorFlow Lite via AI Edge Torch.

\subsection{Evaluation Protocol}
\label{sec:eval_protocol}
Following the challenge guidelines, restoration fidelity is measured using Peak Signal-to-Noise Ratio (PSNR) and Structural Similarity Index (SSIM). Metrics are computed on full-resolution validation images ($2432 \times 3200$) to reflect global structural coherence and the absence of tiling artifacts. In accordance with the official evaluation script, PSNR and SSIM are computed on RGB channels after applying a 1-pixel boundary crop to predicted and ground-truth images. SSIM is computed independently per channel using Gaussian weighting and then averaged.

\subsection{Mobile Runtime Profiling}
On-device profiling is performed on physical MediaTek Dimensity 9500 and Snapdragon 8 Elite handsets using the official AI Benchmark application. Following the benchmark protocol, inference latency is measured in FP16 at Full HD resolution ($1088 \times 1920$). This split-resolution evaluation (Full HD for runtime, full-resolution for PSNR) strictly follows the official challenge requirements, enabling direct runtime comparison across methods while confirming robustness for high-megapixel deployment.
	
\section{Ablation Study}
\label{sec:ablation}

Our ablation study investigates the interaction between architectural complexity and mobile deployment constraints, with a specific focus on NPU-native execution. The study was designed as a boundary-seeking progression to identify design choices that remain compatible with high-resolution mobile inference while defining the physical hardware limits of the target device.

\subsection{Stage 1: Attention-Augmented Baseline}
We first explored an attention-augmented lightweight architecture incorporating spatial and channel attention modules. Although this design was highly compact at approximately 0.34M parameters, it achieved only 31.68 dB PSNR in offline evaluation. Furthermore, the inclusion of non-local attention operators introduced framework-level conversion complexities for high-resolution tensors. This result indicated that extreme lightweighting combined with attention-based modulation lacks the representational capacity required for real-world sensor noise.

\subsection{Stage 2: High-Capacity Teacher}
To establish a quality ceiling, we developed a dense teacher network with 41.6M parameters. This architecture utilized massive convolutional capacity to provide a clean restoration manifold for supervision. The teacher achieved 37.71 dB PSNR, the highest quality reached during development. However, despite successful TFLite conversion, the model exceeded the physical SRAM budget of the NPU, triggering an Out-of-Memory (OOM) failure during 8MP inference. This confirmed that raw capacity is insufficient if the architecture violates device-level memory tiling constraints.

\subsection{Stage 3: Extreme Efficiency Baseline}
To probe the NPU's ``fast-path'' operating range, we evaluated an ultra-lightweight baseline ($f_0=12$) consisting of 345,395 parameters. This model achieved 34.86 dB PSNR with a highly efficient latency of $\sim$26.4 ms. However, the reduced width proved insufficient to recover fine image structures, resulting in unacceptable smoothing of high-frequency textures.

\subsection{Stage 4: Structural Optimization (16-Filter Baseline)}
We increased the student width to 16 base filters (1.96M parameters), establishing a stronger lightweight baseline that remains within the NPU's memory budget. Without distillation, this configuration achieved 36.08 dB PSNR with a stable latency of 46.1 ms. While viable for deployment, a clear 1.63 dB quality gap remained relative to the high-capacity teacher.

\subsection{Stage 5: Proposed Distilled Student}
In the final stage, we applied knowledge distillation to the 1.96M-parameter student using the 41.6M-parameter teacher as supervision. The distilled student reached 37.66 dB PSNR while preserving the 46.1 ms NPU latency. Compared to the teacher, this represents a 21.2$\times$ reduction in parameters while reducing the validation PSNR gap from 1.63 dB in the non-distilled baseline to only 0.05 dB after distillation, thereby recovering 99.8\% of the restoration quality. This proves that distillation into hardware-optimized primitives is the key mechanism for bridging the gap between mobile efficiency and desktop-class fidelity.

\begin{table}[t]
\centering
\caption{Systematic ablation of architectural evolution. PSNR is reported for full-resolution validation ($2432 \times 3200$) to capture high-frequency restoration details. Following standard on-device profiling protocols of the MAI workshop, latency is measured on the NPU at Full HD ($1088 \times 1920$); the fully convolutional design ensures that these FHD metrics serve as a reliable proxy for relative architectural efficiency at $8$MP scales.}
\label{tab:ablation_final}
\resizebox{\columnwidth}{!}{
\begin{tabular}{l l c c c c}
\toprule
\textbf{Stage} & \textbf{Strategy} & \textbf{Params} & \textbf{PSNR (dB)} & \textbf{Latency} & \textbf{Execution Outcome} \\
\midrule
1 & Attention-Augmented & 0.34M & 31.68 & N/A & Compilation Failure \\
2 & \textbf{High-Capacity Teacher} & \textbf{41.6M} & \textbf{37.71} & N/A & \textbf{OOM} \\
3 & 12-Filter Base & 0.35M & 34.86 & 26.4 ms & Valid \\
4 & 16-Filter Base & 1.96M & 36.08 & 46.1 ms & Valid \\
5 & \textbf{Proposed Student (Distilled)} & \textbf{1.96M} & \textbf{37.66} & \textbf{46.1 ms} & \textbf{Valid} \\
\bottomrule
\end{tabular}
}
\end{table}

	\section{Results and Discussion}
\label{sec:results}

\subsection{Quantitative Results}

Our final deployable student achieves 37.66~dB PSNR / 0.9278 SSIM on the full-resolution validation benchmark and 37.58~dB PSNR / 0.9098 SSIM on the held-out test benchmark. These results are obtained with only 1.96M parameters, a storage footprint of 7.52~MB, and a computational cost of 14.13~GMACs at Full HD resolution, while remaining restricted to hardware-friendly operations suitable for stable mobile deployment.

To place our method in the benchmark context, Table~\ref{tab:npu_results} summarizes the challenge results together with the released GPU and NPU runtimes on the two target flagship SoCs: Qualcomm Snapdragon 8 Elite and MediaTek Dimensity 9500. The official challenge ranking is based on average GPU runtime. Since our focus is deployment on dedicated mobile AI accelerators, Table~\ref{tab:npu_results} reports an NPU-oriented reanalysis in which the final score is recomputed by replacing average GPU runtime with average NPU runtime in the official Mobile AI scoring formula.

\begin{table*}[!t]
\centering

\caption{Mobile AI 2026 Image Denoising Challenge results viewed from an NPU-oriented perspective. Methods are ordered by held-out test PSNR in descending order. In accordance with the official challenge requirements, the PSNR and SSIM values were evaluated at full resolution ($2432 \times 3200$), while the runtime values were obtained on Full HD ($1088 \times 1920$) images. PSNR, SSIM, GPU runtime, and NPU runtime values are taken from the organizer-released benchmark table~\cite{mai2026workshop,mai2026report}. The final score reported here is recomputed using the official Mobile AI challenge scoring formula~\cite{ignatov2021mai_denoising}, replacing average GPU runtime with average NPU runtime. The official challenge ranking itself is GPU-based. Methods without valid runtimes on both target NPUs are not assigned an NPU-based final score.}
\label{tab:npu_results}
\resizebox{\textwidth}{!}{
\begin{tabular}{lccccccccc}
\toprule
\textbf{Method / Team} & \textbf{PSNR} $\uparrow$ & \textbf{SSIM} $\uparrow$ & \textbf{Snapdragon GPU (ms)} $\downarrow$ & \textbf{Snapdragon NPU (ms)} $\downarrow$ & \textbf{Dimensity GPU (ms)} $\downarrow$ & \textbf{Dimensity NPU (ms)} $\downarrow$ & \textbf{Avg. GPU (ms)} $\downarrow$ & \textbf{Avg. NPU (ms)} $\downarrow$ & \textbf{Final Score} $\uparrow$ \\
\midrule
\textbf{TeamPAK (ours)} & \textbf{37.58} & 0.9098 & 127 & 46.1 & 138 & 34.0 & 132 & 40.05 & \textbf{139.5} \\
Fanis & 37.48 & 0.9120 & 48.1 & 64.9 & 112 & Failed & 73 & --- & --- \\
TLG & 37.31 & \textbf{0.9129} & \textbf{28} & \textbf{46.0} & \textbf{50.7} & \textbf{18.7} & \textbf{38} & \textbf{32.35} & 118.8 \\
IVCL & 37.28 & 0.9110 & 1961 & 552.0 & 720 & Failed & 1188 & --- & --- \\
Z6 & 36.96 & 0.9001 & 69.3 & 51.6 & 115 & 58.1 & 89 & 54.85 & 43.1 \\
wurbane & 35.22 & 0.8964 & 2292 & 633.0 & 2824 & 524.0 & 2544 & 578.50 & 0.37 \\
Fanis (AIM) & 34.98 & 0.8354 & 182 & 373.0 & 284 & Failed & 227 & --- & --- \\
NormalVision & 33.23 & 0.7767 & 945 & Failed & 1516 & Failed & 1197 & --- & --- \\
\bottomrule
\end{tabular}
}
\vspace{2pt}
\begin{flushleft}
\footnotesize{Based on the organizer-released leaderboard entries, the normalization constant for the recomputed NPU-based score is estimated as $C \approx 7.56 \times 10^{18}$. For our method, the same exported model required 132.0 ms on the integrated mobile GPU under the same Full HD benchmark protocol, corresponding to NPU speedups of $2.86\times$ on the Snapdragon 8 Elite and $3.88\times$ on the Dimensity 9500.}
\end{flushleft}
\end{table*}

Several observations follow from Table~\ref{tab:npu_results}. First, our submission attains the highest held-out test PSNR among the listed methods while recording valid runtimes on both target NPUs. Second, the results indicate that restoration quality and cross-platform NPU compatibility are distinct objectives, and achieving both simultaneously remains non-trivial. Third, among methods with valid runtimes on both NPUs, our model offers a favorable runtime--accuracy trade-off, with runtimes of 46.1 ms on the Snapdragon 8 Elite NPU and 34.0 ms on the Dimensity 9500 NPU.

Although the official challenge ranking is GPU-based and runtime benchmarking follows the official Full HD protocol ($1088 \times 1920$), all reported validation PSNR values are computed on full-resolution images ($2432 \times 3200$) as mandated by the challenge organizers.

On the 50-image validation set, the teacher achieves 37.71 dB PSNR (95\% CI: [36.62, 38.82]) and the distilled student 37.63 dB (95\% CI: [36.53, 38.75]). The mean paired gap is only 0.08 dB (95\% CI: [-0.00, 0.16]), confirming that the distilled model remains very close to the teacher.

\begin{figure}[!t]
    \centering
    \includegraphics[width=\linewidth]{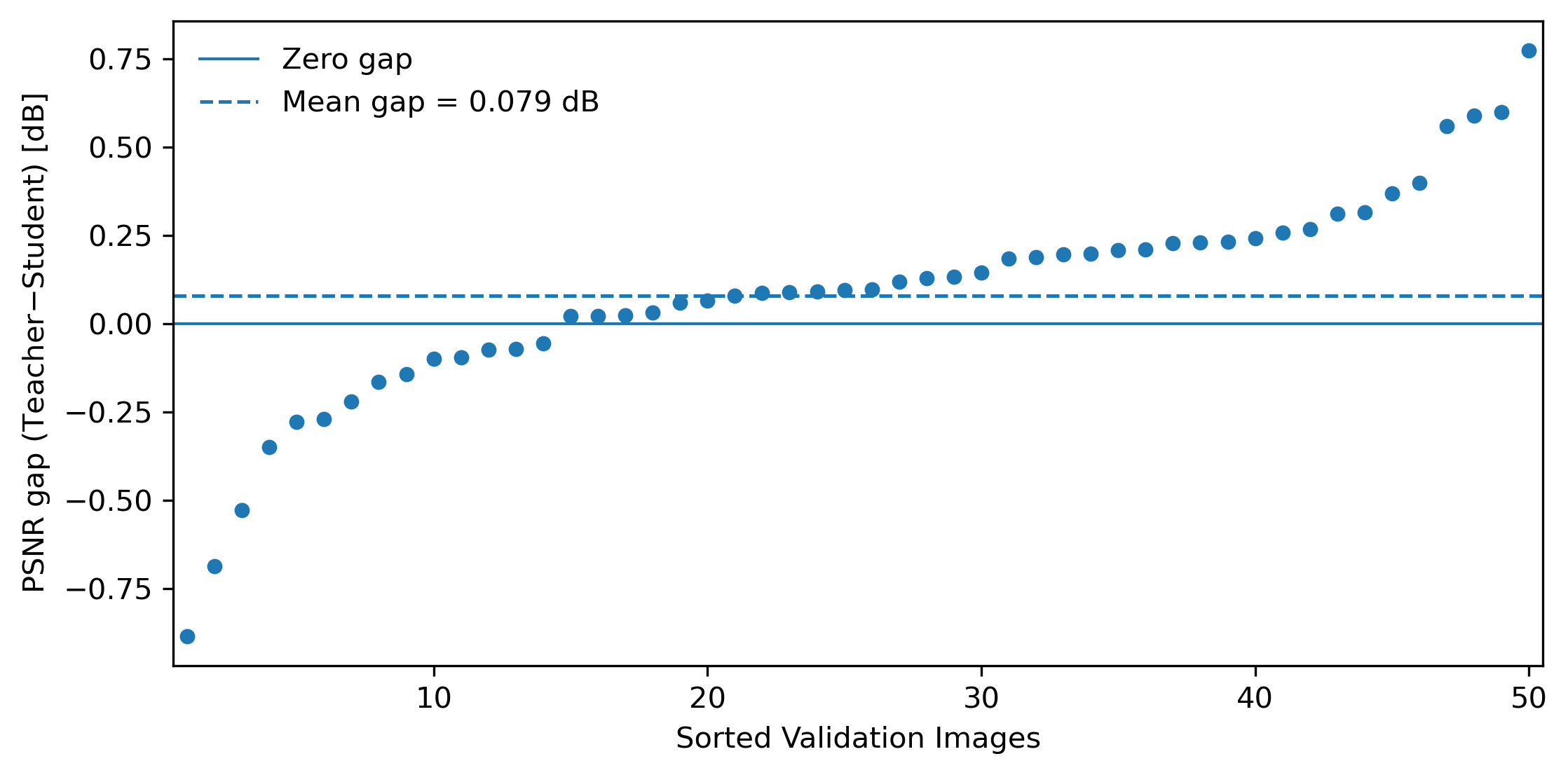}
    \caption{Sorted per-image PSNR gap between the high-capacity teacher and the distilled student on the 50-image validation set. Positive values indicate higher teacher PSNR, while negative values indicate images where the distilled student performs slightly better. Most gaps remain close to zero, consistent with the small mean difference of 0.08 dB.}
    \label{fig:teacher_student_gap}
\end{figure}

Figure~\ref{fig:teacher_student_gap} shows that most per-image PSNR gaps cluster near zero, indicating that the lightweight student recovers nearly all of the teacher's restoration quality. Negative gaps denote images where the distilled student slightly outperforms the teacher.

\subsection{Qualitative Results}

Figure~\ref{fig:teaser} shows a representative qualitative example from the validation set. The proposed lightweight student effectively suppresses visible sensor noise while preserving local structures and visually important textures. In relatively homogeneous regions, the restoration remains clean and stable, whereas in textured regions and along object boundaries the model retains fine-scale details without introducing strong artifacts.

The zoomed crops further illustrate that the student avoids the aggressive oversmoothing often observed in highly compact mobile models. Instead, it produces reconstructions that remain visually natural, with a good balance between denoising strength and detail preservation. This behavior is consistent with the quantitative results and is particularly relevant for practical mobile photography, where edge sharpness, texture fidelity, and visual naturalness are critical to perceived image quality.

\subsection{Inference Inversion on Mobile Accelerators}
We observe an \emph{Inference Inversion} effect: under the Full HD protocol, our model required 132.0~ms on the mobile GPU, versus 46.1~ms (Snapdragon 8 Elite) and 34.0~ms (Dimensity 9500) on the NPU. While mobile GPUs are typically assumed to be the primary acceleration path for restoration, our results show that NPU-native designs can reverse this expectation, making the dedicated NPU the preferred inference target.

Because the proposed student is restricted to hardware-friendly operations---standard $3\times3$ convolutions, ReLU activations, strided downsampling, and nearest-neighbor upsampling followed by convolutional refinement---it maps efficiently onto modern flagship NPUs while avoiding unsupported operators and framework-level fallbacks. Under the official benchmark protocol, this reverses the more traditional mobile-GPU assumption: rather than treating the NPU as a constrained alternative, the dedicated NPU becomes the clearly preferred inference target.

This effect is important beyond the present benchmark. It suggests that for modern mobile imaging systems, operator compatibility and accelerator-native architectural design are central determinants of practical runtime, and that restoration networks may increasingly need to be designed first for NPU execution and only secondarily for general-purpose GPU compatibility.

\subsection{Discussion}
The reported results support the central hypothesis of this work: \emph{hardware compatibility must be treated as a primary design constraint rather than a secondary implementation detail}. The quantitative results, ablation study, and observed inference behavior all reinforce this conclusion. The 41.6M-parameter teacher defines a strong quality upper bound at 37.71~dB PSNR, but exceeds the effective memory budget of the target hardware and fails during high-resolution mobile inference. At the opposite extreme, aggressively lightweight baselines execute efficiently but suffer a substantial loss in restoration fidelity.

The proposed student resolves this tension through hardware-aware distillation. By restricting the architecture to NPU-native primitives and transferring knowledge from the high-capacity teacher, the final 1.96M-parameter model recovers nearly all of the teacher's accuracy while remaining fully deployable on both target devices. Progressive context expansion further improves global structural consistency and full-resolution restoration quality.
As mobile imaging systems increasingly rely on dedicated AI accelerators, the most practically valuable models will jointly optimize restoration quality, operator compatibility, and stable cross-platform execution.

\subsection{Limitations}
This study has several limitations. First, the cross-method runtime values in Table~\ref{tab:npu_results} are taken from the organizer-released benchmark rather than an independently reproduced runtime study. Second, the official runtime protocol is based on Full HD inputs and does not directly represent end-to-end latency for full-resolution 8MP deployment. Third, the conservative operator set ensures deployment stability but may limit more expressive components that could become practical on future mobile NPUs.

\section{Conclusion}
We presented an NPU-aware teacher--student approach for real-world image denoising under practical mobile deployment constraints. The lightweight student, built entirely from hardware-friendly primitives and trained via high-$\alpha$ distillation from a high-capacity teacher, achieves a $21.2\times$ parameter reduction while recovering 99.8\% of the teacher's restoration quality. On the Mobile AI 2026 Image Denoising Challenge, our method achieved 37.66~dB PSNR / 0.9278 SSIM (validation) and 37.58~dB / 0.9098 SSIM (test), with NPU runtimes of 46.1~ms on Snapdragon 8 Elite and 34.0~ms on Dimensity 9500. We further demonstrate an ``Inference Inversion'' effect, where strict adherence to NPU-native operators enables up to $3.88\times$ faster execution than the integrated mobile GPU. These results establish hardware-aware distillation as an effective strategy for unifying high-fidelity denoising with practical cross-platform mobile deployment, motivating future extension to other low-level vision tasks.

\noindent\textbf{Acknowledgments.} This work was partially supported by the Alexander von Humboldt Foundation.

	{
		\small
		\bibliographystyle{ieeenat_fullname}
		\bibliography{bibmain}
	}
	
\end{document}